\documentclass[11pt,a4paper]{llncs}
\usepackage{amsmath}
\usepackage{amssymb}
\usepackage{graphicx}
\usepackage{marvosym}
\usepackage{url}
\usepackage{fancyhdr}
\usepackage{wrapfig}

\usepackage{geometry}
\newcommand{\keywords}[1]{\par\addvspace\baselineskip
\noindent\keywordname\enspace\ignorespaces#1}

\fancypagestyle{firstpage}{\fancyhf{}
}

\begin{document}


\title{\LARGE{Dynamic Multi-Agent Orchestration and Retrieval for Multi-Source Question-Answer Systems using Large Language Models}}


%
%
\author{
    Antony Seabra\inst{1,2} \and 
    Claudio Cavalcante\inst{1,2} \and 
    Joao Nepomuceno\inst{1} \and 
    Lucas Lago\inst{1} \and 
    Nicolaas Ruberg\inst{1} \and 
    Sergio Lifschitz\inst{2}
}

\institute{
    BNDES - Área de Tecnologia da Informação, Rio de Janeiro, Brazil \and
    PUC-Rio - Departamento de Informática, Rio de Janeiro, Brazil
}

\maketitle

\thispagestyle{firstpage}

\begin{abstract}
We propose a methodology that combines several advanced techniques in Large Language Model (LLM) retrieval to support the development of robust, multi-source question-answer systems. This methodology is designed to integrate information from diverse data sources, including unstructured documents (PDFs) and structured databases, through a coordinated multi-agent orchestration and dynamic retrieval approach. Our methodology leverages specialized agents—such as SQL agents, Retrieval-Augmented Generation (RAG) agents, and router agents—that dynamically select the most appropriate retrieval strategy based on the nature of each query. To further improve accuracy and contextual relevance, we employ dynamic prompt engineering, which adapts in real time to query-specific contexts. The methodology’s effectiveness is demonstrated within the domain of Contract Management, where complex queries often require seamless interaction between unstructured and structured data. Our results indicate that this approach enhances response accuracy and relevance, offering a versatile and scalable framework for developing question-answer systems that can operate across various domains and data sources.
\end{abstract}


\section{Introduction}
In recent years, the rapid evolution of Large Language Models (LLMs) has led to significant advancements in the fields of information retrieval and question-answer (Q\&A) systems. These advanced models have proven capable of understanding and generating human-like text, offering new possibilities for retrieving precise and contextually relevant information from diverse sources. However, despite these advancements, challenges remain when integrating data from heterogeneous sources—such as unstructured text documents, structured databases, and real-time APIs—into a single system. Traditional systems often struggle to handle the complexity of retrieving and correlating information across different formats, leading to issues with accuracy and relevance in responses. This gap underscores the need for more sophisticated techniques that can dynamically orchestrate and retrieve information from multiple sources, while maintaining the high accuracy and contextual awareness that LLMs offer.

In many industries, professionals are required to navigate vast volumes of text-based documents while simultaneously accessing structured data from databases or other systems. This process is not only labor-intensive but also time-consuming, as locating specific pieces of information and correlating them across different sources can be very difficult. For instance, in domains like Contract Management, retrieving relevant details from both contract documents and database records can often require manually searching through hundreds of pages and cross-referencing these with structured metadata—an arduous and error-prone task.

To address these challenges, we propose a dynamic multi-agent orchestration and retrieval methodology aimed at improving the accuracy of multi-source Q\&A systems using Large Language Models. By combining advanced retrieval-augmented generation (RAG), text-to-SQL techniques, and dynamic prompt engineering, we enable the system to handle complex queries across heterogeneous data sources, improving response precision without the need to retrain the model. At the heart of this approach lies an agent-based architecture that dynamically orchestrates different retrieval strategies based on the nature of the user query, ensuring optimal data retrieval from multiple sources.

In this paper, we evaluate our approach using the domain of contracts, incorporating qualitative feedback from users who tested the system. Contract Management systems often involve retrieving specific data from contract documents (e.g., penalties, SLAs, deadlines) as well as structured data from databases. While existing systems can handle basic information retrieval tasks, they typically struggle when required to provide detailed answers that integrate information from multiple sources. Our proposed system leverages specialized agents—such as SQL agents, RAG agents, and router agents—to route and execute the queries to the most appropriate source, thereby offering more comprehensive and context-aware responses.

Additionally, we introduce dynamic prompt engineering, which adapts the prompt' instructions in real-time, based on the context of the query, the type of data being retrieved, and the user’s input. This ensures that the language model’s responses are accurate, contextual, and optimized for each query’s specific requirements, whether it’s retrieving information from a structured database or extracting text from an unstructured document.

The paper is organized as follows: Section 2 provides technical background on agents orchestration and retrieval techniques for LLMs, like RAG, text-to-SQL, and Prompt Engineering. Section 3 discusses our methodology and the use of the presented techniques, while Section 4 describes how we evaluated the proposed methodology and the experimentation of the Q\&A application. Finally, Section 5 concludes our study and proposes directions for future research in this field.

\section{Background}
To build an effective multi-source question-answer system, it is essential to leverage several advanced techniques that address the complexities of retrieving and processing information from diverse sources, and orchestrate them using agents. This section explores the foundational technologies that enable the system's core functionality, including Large Language Models (LLMs), which provide the ability to understand and generate natural language; Prompt Engineering, a method used to optimize and guide the behavior of LLMs for specific tasks; Retrieval-Augmented Generation (RAG), which integrates external data into the LLM’s context for more accurate and relevant answers; Text-to-SQL, a technique that translates natural language queries into database commands to retrieve structured data; and Agents, which dynamically orchestrate and route tasks to the most appropriate modules within the system \cite{mialon2023augmented}. Together, these technologies form the backbone of our proposed multi-agent orchestration and retrieval methodology, enabling seamless integration of multiple data sources and improving the overall performance of question-answer systems. 

\subsection{Large Language Models}
Based on the Transformer architecture [Vaswani et al., 2017], Large Language Models (LLMs) have transformed the field of natural language processing (NLP) by enabling machines to generate and comprehend human-like text with remarkable accuracy. These models leverage self-attention mechanisms to evaluate the relevance of different segments of input text, allowing them to capture complex linguistic patterns and relationships more effectively. This architecture enables LLMs to excel at a variety of tasks, from text generation to translation and information retrieval.

The advent of LLMs such as GPT \cite{OpenAI} has significantly advanced the field of Q\&A systems, providing an intuitive interface for retrieving information from diverse data sources. These models can process massive amounts of text data and generate human-like responses, making them suitable for domains requiring natural language understanding. However, while LLMs are powerful, they face limitations such as factual hallucination, outdated knowledge, and challenges in domain-specific expertise \cite{chen2024benchmarking}. To address these limitations, external data sources and retrieval mechanisms, such as Retrieval-Augmented Generation (RAG), have been incorporated into LLM-based systems to provide up-to-date and accurate responses by retrieving relevant information at query time.

\subsection{Retrieval-Augmented Generation (RAG)}
Retrieval-Augmented Generation (RAG) is an advanced technique designed to enhance the performance of LLMs by incorporating external data into the generation process, enabling an extension of knowledge by accessing information that is not part of the LLM's internal knowledge base. While LLMs excel at generating text based on their training, they are limited by the information they contain, which becomes outdated or incomplete over time. RAG addresses this by retrieving relevant documents or data from external sources, such as databases or document repositories, and feeding this information to the LLM as context for generating responses. This ensures that the answers provided are up-to-date and rooted in real-world.

The RAG framework, as described by \cite{gao2023retrieval} and \cite{feng2024retrieval}, operates by embedding both the user's query and chunks of external information into high-dimensional vector spaces. These embeddings allow the system to compare and retrieve the most semantically relevant data from a vectorstore - a database optimized for high-dimensional vectors. Once retrieved, this relevant data is used as additional input to the LLM, ensuring that the generated answer is informed by the latest and most pertinent information available.

A key advantage of RAG is its ability to provide answers that go beyond the internal knowledge of the LLM. This is particularly useful in dynamic environments where the information changes frequently or is too specialized to be captured fully by a pre-trained model. For example, in our multi-source question-answer system, RAG can enable the retrieval of information from both contract documents and databases, ensuring the responses are relevant to the specific user's question.

The chunking strategy used in RAG is critical to its success, as it determines how documents are divided into smaller pieces for embedding and retrieval. By effectively segmenting large documents, RAG ensures that only the most relevant sections are retrieved and fed into the LLM, preventing information overload and improving the precision of the answer. The choice of similarity metrics, such as Cosine or Euclidean distance, plays a significant role in determining which chunks are selected for retrieval \cite{gao2023retrieval}. In RAG, the chunking strategy is important because it directly influences the quality of the retrieved information. A well-designed chunk generation ensures that the information is cohesive and semantically complete, capturing its essence. Several chunking options can be applied depending on the structure and type of data. For instance, one common approach is to divide text into chunks based on a specific number of tokens, often with an overlap parameter to ensure continuity between chunks. This overlap helps maintain context, especially in lengthy documents where relevant information may span across multiple chunks. Another approach, particularly suited for uniform documents, involves chunking based on specific sections or headers within the document, such as dividing contracts by clauses or legal sections. This ensures that each chunk represents a self-contained, semantically meaningful portion of the text. The choice of chunking method plays a crucial role in determining the precision of retrieval, as it helps balance the trade-off between capturing full context and maintaining relevance in the retrieved information.

While RAG is highly effective in bridging the gap between static knowledge in LLMs and real-time data, it also presents challenges, particularly when the retrieved chunks are semantically similar but not relevant to the query. This issue often arises in scenarios involving structured documents, such as contracts, where different sections may contain similar language but vastly different meanings. The difference between similarity and relevance is one of the biggest challenges faced by retrieval systems, especially when working with large language models. While similarity measures help in identifying content that closely matches the query in terms of wording or context, it does not always guarantee that the retrieved information is truly relevant to the user’s intent. This challenge often requires additional filtering or refinement techniques to ensure that retrieved results are not just similar but also meaningful and aligned with the specific needs of the query.

\subsection{Text-to-SQL}
Text-to-SQL is a powerful technique that bridges the gap between natural language queries and relational database systems by converting user inputs in plain text into executable SQL commands. This allows users to retrieve precise, structured information from databases without needing to understand SQL syntax \cite{liu2023comprehensive}. By leveraging the capabilities of LLMs, Text-to-SQL systems can parse and interpret natural language questions and map them to the appropriate database schema, significantly improving the accessibility of data for non-expert users.

A key advantage of Text-to-SQL systems is their ability to handle complex database queries while shielding users from the intricacies of database schemas and SQL commands. For instance, as discussed by \cite{pinheiro2023construction}, LLMs can be used to
construct natural language database (conversational)
interfaces. They do this detecting entities, mapping them to corresponding tables and columns, and generating syntactically correct SQL queries based on the database structure. This approach is particularly useful in domains where the underlying data is stored in complex databases, such as contract management systems or healthcare databases, where queries may involve multiple tables and relationships.

According to \cite{sbbdContrato360}, the main distinction between RAG and text-to-SQL techniques lies in their approach to retrieving information. RAG focuses on retrieving text segments from a vectorstore that are semantically similar to the user’s question, and it uses these segments to generate a coherent and contextually appropriate answer. This approach is well-suited for questions where the answer can be synthesized from existing unstructured text. However, it may not always provide the precise information expected if the answer cannot be directly inferred from the retrieved text segments. On the other hand, Text-to-SQL translates natural language queries into SQL commands, as demonstrated in \cite{pinheiro2023construction}, which are then executed against a structured database to return exact data matches. This ensures that when the text-to-SQL translation is accurate, the user receives a highly specific, structured answer derived directly from the relevant database fields.

Therefore, while RAG operates on the principle of textual similarity and utilizes generative capabilities to synthesize responses from retrieved text, Text-to-SQL provides a more direct and precise mechanism for data retrieval. By translating natural language queries into executable SQL commands, Text-to-SQL allows for exact matches based on the user’s intent, retrieving highly specific information directly from structured databases. This makes Text-to-SQL particularly effective for data investigations where precise, query-based access to relational data is crucial, such as financial reports, contract details, or inventory systems. Unlike RAG, which depends on finding semantically similar text, Text-to-SQL guarantees an exact match from database fields, ensuring that the user receives accurate, factual answers without ambiguity. As a result, it is a valuable tool in scenarios where precision and structure are paramount, complementing the generative and flexible nature of RAG for a more comprehensive information retrieval system.

\subsection{Prompt Engineering}
Prompt Engineering guides and optimizes the behavior of LLMs using direct instructions, ensuring that the generated responses align with the user’s intent. By carefully crafting the input prompt, developers can influence not only the content of the response but also its tone, format, and level of detail \cite{OpenAIprompt}. This technique becomes especially important in multi-source question-answer systems, where prompts must clearly define the task at hand, instruct the model on how to handle various data types, and help the model distinguish between relevant and irrelevant information.

A well-constructed prompt can dramatically improve the accuracy and relevance of the answers generated by LLMs. Engineers can even outline the script for a response, specifying the desired style and format for the LLM response, as stated by \cite{white2023prompt} and \cite{giray2023prompt}. For instance, when querying contract details from unstructured documents or structured databases, the prompt can explicitly instruct the model to only consider the relevant sections of the contract or to retrieve specific details, such as deadlines or penalties. By embedding instructions, context, and constraints into the prompt, it becomes possible to guide the LLM toward more focused and precise outputs. This is especially useful in domains where accuracy is critical, such as legal, healthcare, or finance, where responses must adhere to specific guidelines or regulatory frameworks.

According to \cite{wang2023unleashing}, prompts provide guidance to ensure that the model generates responses that are aligned with the user’s intent. For example, the prompt can be designed to include contextual information that helps the LLM understand the role of the user or the nature of the query. In a contract management system, a prompt might instruct the model to retrieve details about penalty clauses or contractual obligations, with a directive such as: “Extract and summarize any penalty-related clauses from the contract document, focusing on late delivery penalties.” Additionally, the prompt might include a role-specific context like: “You are a contract management assistant tasked with summarizing the key contractual obligations of the supplier.” These instructions help the model generate responses that are not only factually accurate but also aligned with the user’s expectations and the context in which the information is needed.

Moreover, prompt engineering can address ambiguity and reduce the risk of factual hallucination, a common issue where LLMs generate responses that sound plausible but are not grounded in factual data. By explicitly defining the scope of the query in the prompt and instructing the model to refer only to external data sources or documents, the accuracy of responses is improved. For example, by using instructions like “Do not use prior knowledge”, prompt engineering can restrict the LLM’s answer to specific sources, thereby reducing the risk of factual hallucinations and ensuring that responses are grounded in the desired information.

Recent studies have begun to explore the synergistic integration of these techniques with LLMs to create more sophisticated Q\&A systems. For example, \cite{jeong2023study} reinforces the importance of using Prompt Engineering with RAG to improve the retrieval of relevant documents, which are then used to generate both contextually relevant and information-rich answers. Similarly, \cite{gao2023text} explores the integration of Text-to-SQL with Prompt Engineering to enhance the model's ability to interact directly with relational databases, thereby expanding the scope of queries that can be answered accurately.

\subsection{Agents}
In the context of question-answer systems, agents play a pivotal role in orchestrating complex workflows and dynamically routing queries to the most appropriate processing path. Unlike traditional, linear retrieval systems, agent-based architectures introduce a level of flexibility and intelligence, enabling systems to handle multi-faceted queries with varying data sources and retrieval requirements. By incorporating agents, a question-answer system can dynamically adapt to the user’s query, selecting different strategies based on the type of information being requested and the nature of the data sources. 

In an agent-based framework, specialized agents can be designed to handle specific tasks, each tailored to different aspects of information retrieval. For example, a Router Agent can serve as the system’s primary decision-maker, analyzing each query upon receipt and deciding on the optimal retrieval strategy. The router agent typically uses rules, such as regular expressions or other pattern-matching methods, to interpret the query structure and identify key indicators that suggest which retrieval path should be followed. For instance, if the query pertains to a specific clause or passage within a document, the router agent can direct the query to a Retrieval-Augmented Generation (RAG) Agent, which is optimized for unstructured text data. Alternatively, if the query requires precise, structured information, such as dates, financial figures, or other exact data, the router agent might direct it to a SQL Agent, which uses Text-to-SQL translation to interact with structured databases. These agents leverage recent advancements in AI, such as RAG and Text-to-SQL, to perform more complex and contextually aware tasks \cite{lewis2020retrieval}.

Each specialized agent in this framework brings unique capabilities that contribute to the overall effectiveness of the system. The RAG Agent operates by retrieving relevant text chunks from a vectorstore based on semantic similarity and then integrating these chunks into the language model’s context. This allows the system to handle complex, interpretive questions that benefit from a nuanced understanding of unstructured text. Meanwhile, the SQL Agent translates natural language queries into SQL commands, enabling the system to retrieve precise, structured data directly from a database. This approach is particularly useful for answering fact-based questions that require a high degree of specificity, ensuring that the response reflects up-to-date information directly from the database. As outlined in \cite{singh2024enhancing}, agent workflows allow LLMs to operate more dynamically by incorporating specialized agents that manage task routing, execution, and optimization. These agents serve as intelligent intermediaries, directing specific tasks—such as data retrieval, reasoning, or response generation—to the most suitable components within the system. One of the most important ones in place are the Router Agents, as they are the decision-makers of the system. When a user poses a query, the router agent analyzes the input and decides the best path forward. 

Through the use of such agents, the question-answer system gains the ability to make intelligent decisions about query handling and data retrieval. This agent-based orchestration allows the system to seamlessly blend information from both structured and unstructured sources, improving the relevance and accuracy of the answers provided. Moreover, agents allow for modular expansion, meaning that new agents can be added to handle specific types of data or tasks, enhancing the system’s scalability and adaptability to diverse domains. According to \cite{jin2024llms}, applying LLMs to text-to-database management and query optimization is also a novel research direction in natural language to code generation task.

In sum, agents empower question-answer systems with the capability to route queries intelligently, select the best processing pathway, and ensure that the response leverages the most suitable data sources. This orchestration enables the system to handle complex, multi-source queries more effectively, providing responses that are both contextually rich and highly relevant to the user’s intent.

\section{Our Methodology}
In designing our multi-source question-answer methodology, we employ a combination of advanced techniques to access diverse data sources and provide accurate responses tailored to the query and the specific source of information, integrating Retrieval-Augmented Generation (RAG), Text-to-SQL, Dynamic Prompt Engineering, and Agent-based orchestration to effectively manage the complexities of interacting with both structured and unstructured data sources. Each component plays a critical role in handling various aspects of information retrieval, ensuring that the system can dynamically adapt to the requirements of each query.

\begin{figure*}[ht]
\centering
\includegraphics[width=\textwidth]{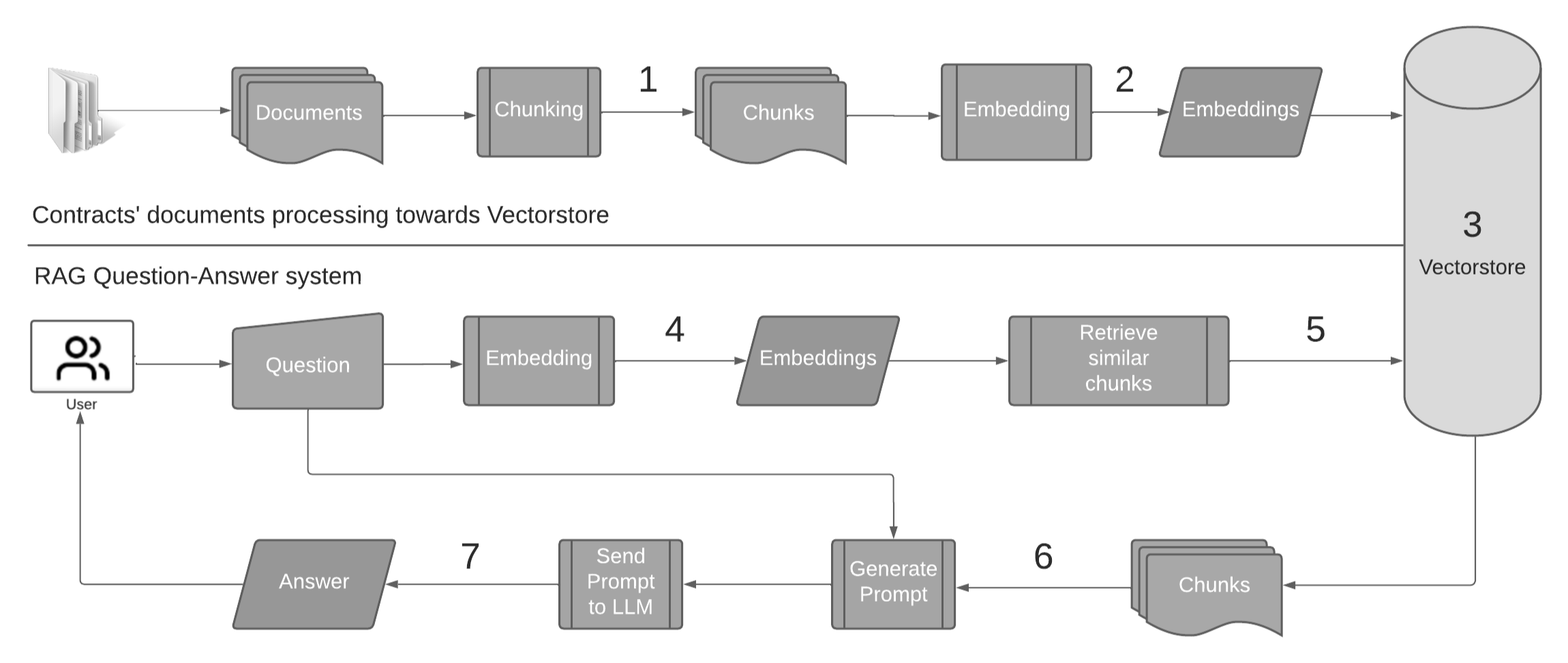}
\caption{Retrieval-Augmented Generation.}
\label{fig:rag}
\end{figure*}

RAG enables the retrieval of relevant information from large volumes of unstructured text, while Text-to-SQL facilitates precise access to structured data within relational databases. Dynamic Prompt Engineering customizes the query context, ensuring that responses are tailored to user intent, and Agent-based orchestration coordinates these techniques, directing queries to the appropriate modules and managing workflows seamlessly. In this section, we detail the approaches and challenges associated with implementing each of these techniques, along with the strategies we used to optimize their integration.

Our methodology was implemented and tested in a real-world project called Contrato360 \cite{sbbdContrato360}, a question-answer system designed specifically for Contract Management. Contrato360 leverages the combined techniques of Retrieval-Augmented Generation (RAG), Text-to-SQL, Dynamic Prompt Engineering, and Agent-based orchestration to address the unique challenges of navigating and retrieving information from complex contract documents and structured databases. By integrating these advanced methods, Contrato360 enables users to efficiently query contract-related data, such as penalty clauses, deadlines, and contractual obligations, across diverse sources. This practical application demonstrates the effectiveness of our methodology in a domain where accuracy, relevance, and contextual understanding are critical.

\subsection{Applying RAG}
According to \cite{sbbdContrato360}, the first step when applying RAG involves (1) reading the textual content of the PDF documents into manageable (\textit{chunks}), which are then (2) transformed into high-dimensional vectors (\textit{embedding}). The text in vector format captures the semantic properties of the text, a format that can have 1536 dimensions or more. These \textit{embeddings} (vectors) are stored in a \textit{vectorstore} (3), a database specialized in high-dimensional vectors. The vector store allows efficient querying of vectors through their similarities, using the distance for comparison (whether \textit{Manhatan}, Euclidean or cosine). Once the similarity metric is established, the query is \textit{embedded} in the same vector space (4); this allows a direct comparison between the vectorized query and the vectors of the stored chunks, retrieving the most similar chunks (5), which are then transparently integrated into the LLM context to generate a \textit{prompt} (6). The \textit{prompt} is then composed of the question, the texts retrieved from the \textit{vectorstore}, the specific instructions and, optionally, the \textit{chat} history, all sent to the LLM which generates the final response (7).

\subsubsection{Chunking strategy}
One of the first decisions to be made when applying RAG is to choose the best strategy to segment the document, that is, how to perform the \textit{chunking} of the PDF files. A common \textit{chunking} strategy involves segmenting documents based on a specific number of \textit{tokens} and an overlap (\textit{overlap}). This is useful when dealing with sequential texts where it is important to maintain the continuity of the context between the \textit{chunks}.

There is a common type of document with well-defined sections; contracts are a prime example. The have a standardized textual structure, organized into contractual sections. Therefore, sections with the same numbering or in the same vicinity describe the same contractual aspect, that is, they have similar semantics. For example, in the first section of contract documents, we always find the object of the contract. In this scenario, we can assume that the best \textit{chunking} strategy is to separate the \textit{chunks} by section of the document. In this case, the \textit{overlap} between the \textit{chunks} occurs by section, since the questions will be answered by information contained in the section itself or in previous or subsequent sections. For the contract page in the example in Figure~\ref{fig:contracts}, we would have a \textit{chunk} for the section on the object of the contract, another \textit{chunk} for the section on the term of the contract, that is, a \textit{chunk} for each clause of the contract and its surroundings. This approach ensures that each snippet represents a semantic unit, making retrievals more accurate and aligned with queries.

Using predefined sections as the boundaries for \textit{chunks} enhances the relevance of responses within a single contract. However, this approach presents two main challenges: (1) within a single document, when a term appears repeatedly, it can be difficult to identify the specific chunk that answers a question; and (2) as the number of documents increases, accurately selecting the correct document to address becomes more challenging for the system. In the Contract Management domain, consider a scenario where the user asks, "Who is the contract manager of contract number 123/2024?". This query is intended to retrieve the specific name of the contract manager for the given contract. However, the term “contract manager” can appear in various clauses of the contract document, often in sections that do not contain the name of the actual manager but refer to responsibilities or general rules related to contract management. For instance, multiple clauses across different sections of the contract might mention the term "contract manager" in contexts like assigning responsibilities, explaining the duties of a manager, or defining roles in contract supervision. Even though these clauses contain the term "contract manager," they do not answer the user's question, which is specifically asking for the name of the contract manager for contract 123/2024.

\begin{wrapfigure}{l}{0.6\textwidth}
\begin{center}
\includegraphics[width=1\linewidth]
{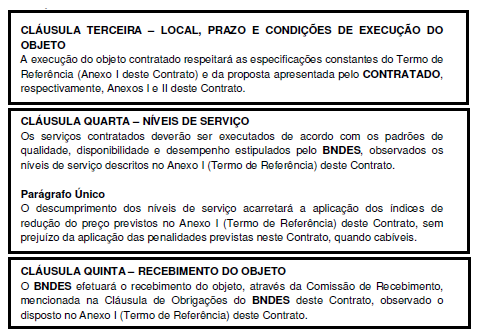}
\end{center}
\caption{Chunking based on Contract's clauses} 
\label{fig:matedados}
\end{wrapfigure}

Due to the similarity between the query and these irrelevant sections, the Retrieval-Augmented Generation (RAG) system may retrieve a chunk from one of these irrelevant clauses that does not actually contain the required name. For example, instead of retrieving the clause that explicitly names the contract manager, the system might retrieve a clause that discusses the general duties of a contract manager. This happens because the chunk embedding for a clause about the role or responsibilities of the manager may be semantically similar to the query, even though it lacks the specific information requested. In this case, the chunk retrieved is related to the term "contract manager" but does not include the answer the user expects. As a result, the system could return an incorrect response, such as a general description of the role of a contract manager, rather than identifying the actual manager for contract 123/2024. 
This illustrates the challenge of relying solely on textual similarity in chunk retrieval, as it can lead to the retrieval of information that is similar to the query in wording but not relevant to the specific context of the user's question. To mitigate this, additional filtering mechanisms, such as metadata checks or contract-specific identifiers, are required to ensure that the system retrieves the most contextually appropriate information from the correct contract section.

To overcome this issue, several strategies can be applied. One approach is to add metadata to the \textit{chunks} and, when accessing the \textit{vectorstore}, use this metadata to filter the information returned. This method improves the relevance of the retrieved texts by narrowing the search to only those chunks that match specific metadata criteria. Figure~\ref{fig
} displays the most relevant metadata attributes for the contracts: \textit{source}, \textit{contract}, and \textit{clause}. Here, \textit{source} represents the name of the contract's PDF file, \textit{contract} refers to the contract number, and \textit{clause} indicates the section title. For instance, when querying, "Who is the contract manager of contract 123/2024?" the system first filters for chunks that belong to contract number 123/2024 and clauses related to the contract manager. Once these chunks are filtered, a similarity calculation is applied to identify the most relevant text segments, which are then sent to the LLM to generate the final response.

\begin{wrapfigure}{l}{0.4\textwidth}
\begin{center}
\includegraphics[width=1\linewidth]
{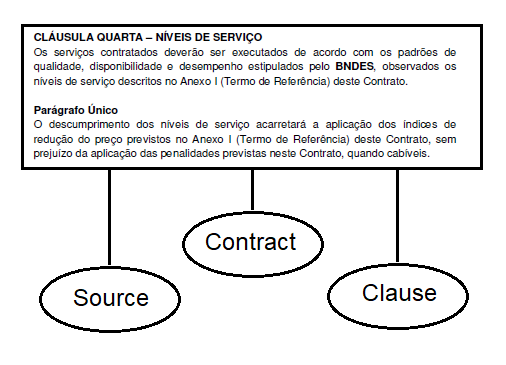}
\end{center}
\caption{Chunk's metadata} 
\label{fig:matedados}
\end{wrapfigure}

\subsubsection{Embeddings models}
Embedding models are a cornerstone of modern NLP tasks and plays in importante role in our methodology. These models transform words, sentences, or even entire documents into high-dimensional vectors, or embeddings, and the key advantage of embeddings is that they enable more nuanced and semantically aware operations on text data, such as similarity comparisons and clustering. By embedding both the query and the text chunks in the same vector space, the system can measure how close they are to each other in meaning, ensuring that relevant information is retrieved even when it is not an exact keyword match. 

Selecting the right embedding model depends on several factors related to the specific needs of a task, including the type of data, the complexity of the queries, and the computational resources available. Pretrained Models, such as BERT or GPT, are trained on vast amounts of general-purpose text data and are  ideal for general tasks where the text spans multiple domains or where high-quality embeddings are required without the need for domain-specific customization. By contrast, custom models work better in specialized fields like legal or medical domains, as they can be beneficial to train an embedding model on a domain-specific corpus. This can help the model better capture the unique terminology and context of that field.

With respect to the vectors' dimensionality, embedding vectors can range in dimensionality depending on the model and the task. For instance, models like GloVe or Word2Vec often produce lower-dimensional embeddings (e.g., 300 dimensions), whereas modern transformer-based models like BERT and GPT can produce embeddings with 768 or more dimensions. Higher-dimensional embeddings typically capture more information and are better for complex tasks like Q\&A systems or semantic search, but they also require more computational resources and storage. Lower-dimensional embeddings are computationally cheaper and faster but may not capture as much nuance, making them better suited for simpler tasks like keyword matching. If precision and detailed contextual understanding are important, high-dimensional embeddings are the better choice. For simpler or resource-constrained tasks, lower-dimensional embeddings may suffice.

In designing our multi-source Q\&A methodology, we carefully evaluated various options for embedding models and vector dimensionality to optimize the system’s performance. After considering several alternatives, we selected text-davinci-002, a model from OpenAI’s GPT-3.5 family, along with embeddings with 1536 dimensions to strike a balance between accuracy, context understanding, and computational efficiency. One of the main advantages of text-davinci-002 is its ability to handle long sequences of text while maintaining a clear understanding of the context. This is essential when dealing with lengthy  documents where information can be dispersed across various sections. The model can track the user’s query context and dynamically retrieve or generate responses that are coherent and relevant to the query. With 1536 dimensions, the embeddings can better represent the complex relationships between terms in the text, especially in documents where meaning often depends on subtle distinctions in wording. This is particularly useful in distinguishing between similar but contextually different terms, such as contract manager vs. contract supervisor, ensuring that the system retrieves the most relevant chunks.

\subsubsection{Vectorstore}
The need to store and query high-dimensional vectors efficiently has led to the development of specialized vector databases, also known as vectorstores. These databases allow for the storage and retrieval of vector embeddings, making it possible to perform similarity searches - a key operation in tasks such as Retrieval-Augmented Generation (RAG) and semantic search. Unlike traditional databases that are optimized for structured, tabular data, vector databases are designed to handle embeddings generated by models like text-davinci-002, which represent semantic relationships in high-dimensional space.

When choosing the right vector database for a project, several factors come into play, including scalability, ease of use, latency, and integration with machine learning models. In our work, we evaluated three popular vector databases: Pinecone, Weaviate, and ChromaDB. Pinecone is a cloud-native vector database that excels in providing a fully managed service for high-performance similarity search. Weaviate is an open-source vector database that provides a highly flexible, schema-based approach to storing and querying vectors alongside structured metadata. ChromaDB is an open-source, lightweight vector database that focuses on simplicity and tight integration with machine learning workflows, making it ideal for embedding-based retrieval tasks in research and smaller projects. Our choice was the last one, specially because ChromaDB is easy to set up and integrate into a project without requiring extensive configuration or overhead. Given that our system is heavily Python-based, ChromaDB’s Python-first design allowed us to quickly embed it into our machine learning pipelines. This streamlined our development process, enabling rapid iteration and testing, which was especially important in the early stages of system design. Also, by using ChromaDB, we can directly connect our text-davinci-002 embeddings with the vectorstore, enabling efficient similarity searches and accurate retrieval of contextually relevant information.

\subsubsection{Similarity searches}
Similarity search is a fundamental operation in tasks that involve comparing vector embeddings to find data points that are semantically or contextually similar. This technique is widely used in fields such as information retrieval, recommendation systems, question-answering systems, and semantic search. The core of similarity search lies in the ability to measure how “close” two vectors are to each other in a high-dimensional space. Several distance metrics are commonly used to quantify this similarity, each with its own strengths and weaknesses depending on the nature of the data and the task. Three of the most popular algorithms for similarity searches include Cosine similarity, Euclidean distance, and Manhattan distance. Each method has a unique approach to measuring how similar two vectors are, and the choice of algorithm can significantly impact the performance and accuracy of a similarity-based system.

Cosine similarity measures the cosine of the angle between two vectors in a multi-dimensional space. It evaluates how “aligned” the two vectors are rather than how far apart they are. The cosine similarity value ranges from -1 to 1, where 1 indicates that the vectors are perfectly aligned (very similar), 0 means that the vectors are orthogonal (completely dissimilar), and -1 indicates that the vectors point in opposite directions. Cosine similarity is often used in text-based applications, where the magnitude of the vector is not as important as the direction. Euclidean distance is the most common metric for measuring the straight-line distance between two points (or vectors) in a multi-dimensional space. It calculates the “as-the-crow-flies” distance between two vectors, treating each dimension as an axis in a Cartesian plane. Euclidean distance is widely used in geometric tasks or where the actual distance between points matters. Manhattan distance, also known as L1 distance or taxicab distance, measures the sum of the absolute differences between the corresponding coordinates of two vectors. Instead of measuring the direct straight-line distance (as in Euclidean), Manhattan distance measures how far one would have to travel along the axes of the space.

In our work, we chose cosine similarity for its ability to prioritize semantic alignment between query embeddings and document embeddings. Its strength in handling high-dimensional data, minimizing the influence of vector magnitude, and focusing on the directionality of vectors makes it the ideal choice for our Q\&A system methodology. Cosine similarity is widely recognized as one of the best similarity measures for text-based applications, especially when using vector embeddings generated from NLP models like text-davinci-002. Since our system heavily relies on textual data, cosine similarity was the natural choice for ensuring that user queries are matched with the most relevant sections of the text, even if the exact phrasing differs. Whether we are retrieving specific sections in documents or providing general answers based on lenghty documents, cosine similarity ensures that the system is aligned with the semantic intent of the query.

\subsection{Using structured data}
In order to improve our question-answer system methodology, we explored two distinct approaches to integrate data from structured databases effectively. The first approach involved extracting data directly from the database, transforming it into text, and embedding this text into vector representations stored in the same vectorstore as our document-based embeddings. This method allowed us to convert structured data into a more flexible, text-based format, enabling semantic similarity searches alongside the unstructured text from contract documents. By embedding database information in this way, we created a unified search space where both structured and unstructured data could be queried with the same similarity-based techniques. This approach offered the advantage of simplicity, as it enabled direct integration of database information into our existing RAG framework, ensuring that queries could retrieve relevant data without needing to connect to the database during runtime.

The second approach we implemented involved a Text-to-SQL method, where natural language questions are dynamically translated into SQL queries. In this setup, the system interprets the user’s query, converts it into a structured SQL command, and then submits it to the database for execution. The Text-to-SQL approach allows for precise data retrieval by directly querying the database, which is particularly beneficial for questions requiring exact, up-to-date values, such as specific dates, contract numbers, or quantitative information. Unlike the first approach, this method does not rely on pre-embedded representations; instead, it provides real-time access to structured data, ensuring that answers are accurate and reflect the current database state.

Each approach has its advantages. Embedding database data alongside unstructured text provides a unified search experience and reduces dependence on real-time database access. In contrast, the Text-to-SQL approach supports direct and precise querying, making it ideal for cases where exact values are necessary. Together, these approaches allow the system to leverage the strengths of both pre-embedded and dynamic querying, enhancing its versatility in handling a wide range of user queries.

\subsection{Agents}
Agents are central to the functionality and adaptability of our multi-source question-answer system, enabling it to handle diverse query types efficiently. By leveraging specialized agents, the system dynamically routes each query to the most suitable processing pathway, ensuring that user questions are handled with precision and contextual relevance. In our architecture, the Router Agent serves as the primary decision-maker, evaluating each incoming query and directing it to the appropriate agent based on predefined criteria.

\begin{figure*}[ht]
\centering
\includegraphics[width=1\textwidth]
{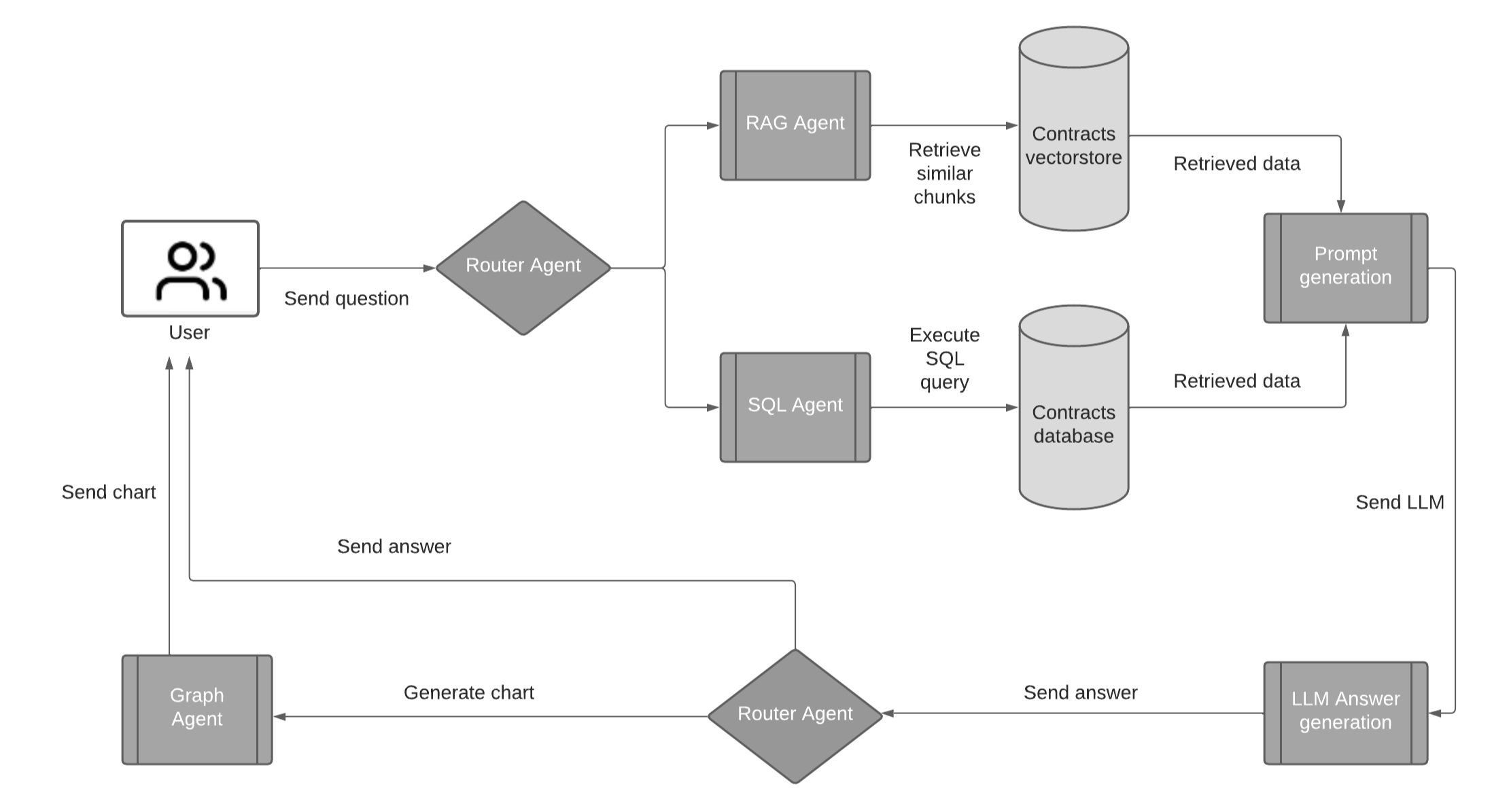}
\caption{Agents Architecture.}
\label{fig:agents}
\end{figure*}

The Router Agent uses regular expressions to identify keywords, patterns, or structures within the query. If the query is specific to a clause within a contract, the Router Agent recognizes this pattern and assigns the query to the RAG Agent. The RAG Agent is optimized for handling unstructured text data, retrieving relevant text chunks from the vectorstore. By focusing on textual similarity, the RAG Agent retrieves semantically aligned information and generates responses that incorporate precise, contextually relevant excerpts from the documents, addressing the specifics of the the user’s question.

Conversely, if the Router Agent detects that the question involves broader contract information, such as dates, financial details, or other exact values, it directs the query to the SQL Agent. The SQL Agent translates the natural language question into a structured SQL query, which is then executed against the database to retrieve exact data. This approach is particularly effective for queries requiring precise, structured responses, ensuring that the system provides accurate and up-to-date information directly from the database.

This dynamic agent-based architecture enables our system to handle both unstructured and structured data seamlessly. The Router Agent’s decision-making process allows the system to optimize query processing based on the context and specific needs of each query. By directing contract-specific questions to the RAG Agent and structured data queries to the SQL Agent, the Router Agent ensures that user questions are handled efficiently, providing relevant answers whether they require interpretive text or exact data values. This modular design not only improves response accuracy but also enhances the system’s flexibility in adapting to a wide range of contract-related queries.

\subsection{Dynamic Prompt Engineering}
\label{sec:applicandoPrompt}
In our work, we use Prompt Engineering to enhance the accuracy of generated answers, guiding the LLM's behavior to ensure responses are contextually relevant and tailored to the user’s needs. We utilize dynamic prompts that adapt according to the specific agent handling the query. By tailoring prompts to each agent, we ensure that every query receives an optimal response, whether it involves unstructured text, structured data, or visual representation.

For instance, when a query is managed by the RAG Agent, the prompt is dynamically constructed to include relevant contextual instructions that guide the LLM in synthesizing information from text chunks retrieved from the vectorstore. This allows the model to draw on semantically similar text embeddings while aligning with the specific details of the user’s question. For queries handled by the SQL Agent, the prompt is designed to capture the user’s intent in a structured format, translating natural language into a precise SQL command that retrieves exact information from the database. This approach ensures that the LLM responds with high accuracy when the query requires structured data or specific values, such as contract dates or financial figures.

Additionally, we developed a Graph Agent to enrich responses with visual information, especially when dealing with tabular or quantitative data. When the LLM’s output includes values suited for visual representation, the Graph Agent dynamically prompts the model to interpret this data and present it as a bar graph. This feature is particularly useful for queries that involve comparisons, trends, or grouped data, providing users with clear, visual insights in addition to textual explanations. By incorporating graph-based responses, our system enhances user understanding, making complex data more accessible and interpretable.

To illustrate, consider a Contract Management Q\&A system where dynamic prompts are applied in real-time. If a user asks the RAG Agent, "What are the responsibilities of the contract manager for contract 123/2024?", the prompt is constructed as follows: "Retrieve relevant sections from contract 123/2024 that detail the role and responsibilities of the contract manager. Use information from clauses specifying contract management tasks." This tailored prompt focuses the LLM on extracting specific responsibilities, enhancing relevance and accuracy.

Alternatively, if a user asks the SQL Agent, "Who are the managers of contracts that we have with IBM?", the prompt is dynamically structured to interpret this query in SQL form and guide the LLM to provide a table in its response. The prompt is transformed into the following instructions: "Retrieve a list of active contracts with IBM, displaying each contract manager’s name in a table format."

Through dynamic prompt engineering, our system adapts the behavior of the LLM based on the specific needs of each agent, whether generating text from retrieved information, executing SQL queries, or displaying data visually. This approach ensures that responses are contextually accurate, actionable, and user-friendly, enhancing the overall functionality and versatility of the question-answer system.

\section{Evaluation}
The architecture depicted in the figure represents the implementation of our multi-source question-answer methodology, combining structured and unstructured data from contracts. The system is built using a modular approach, where each component plays a critical role in the data retrieval and response generation process. At the core of the architecture is the User Interface, built with Streamlit, as shown in figure \ref{fig:interface}, which allows users to input their queries and view responses in a user-friendly interface. Users can submit both broad questions or specific contract-related queries, which are then processed by the backend system.

\begin{figure*}[ht]
\centering
\includegraphics[width=.7\textwidth]
{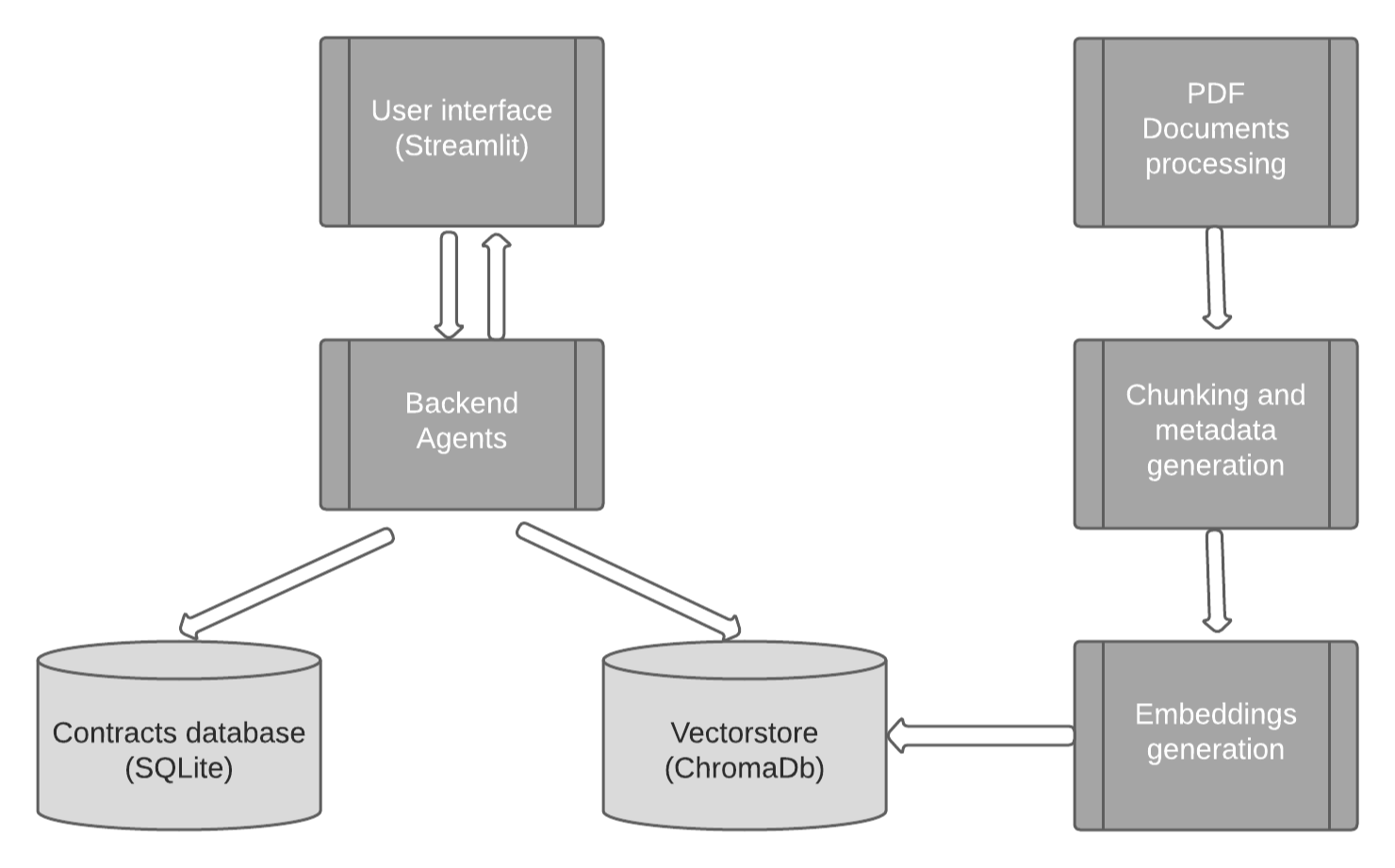}
\caption{Application architecture.}
\label{fig:architecture}
\end{figure*}

The Backend Agents act as the decision-making layer of the system, handling queries based on their type and content. These agents include the Router Agent, which determines whether to route the query to the RAG Agent (for unstructured text retrieval) or the SQL Agent (for structured data queries using Text-to-SQL). The agents communicate bidirectionally with the user interface, allowing for interactive feedback during the query resolution process.

For the unstructured data flow, contract documents in PDF format undergo processing in the PDF Documents Processing component. This involves extracting text and metadata from the documents, which is then passed to the Chunking and Metadata Generation module. This module divides the documents into manageable chunks, enriching them with metadata for easier retrieval. These chunks are further processed through the Embeddings Generation component, where each chunk is transformed into a high-dimensional vector representation using an embedding model. These embeddings are stored in the Vectorstore (implemented using ChromaDB) for efficient similarity search during retrieval.

On the structured data side, the Contracts Database (implemented using SQLite) stores relevant contract data such as specific terms, clauses, dates, and financial information. When a query requires precise data retrieval, such as asking for contract values or deadlines, the SQL Agent retrieves the necessary information directly from this database.

By integrating both the vectorstore and structured database, the Backend Agents can provide comprehensive answers to user queries, dynamically choosing the most appropriate data source based on the type of question. This hybrid approach ensures that the system can handle both semantically complex queries and direct database queries, offering flexible and accurate responses.

\begin{figure}[ht]
\centering
\fbox{\includegraphics[height=.8\linewidth,width=.8\linewidth]{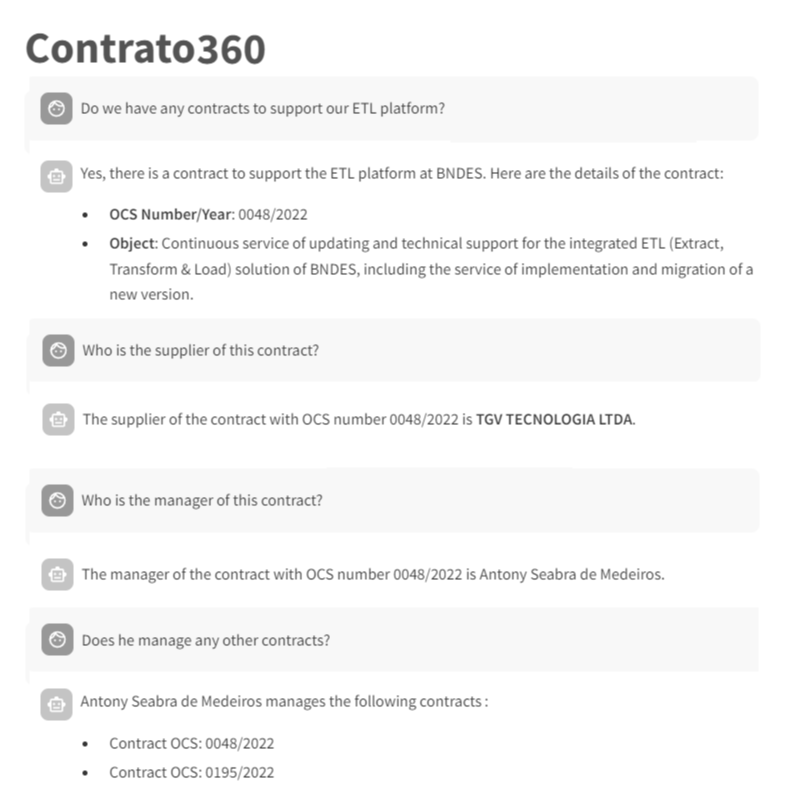}}
\caption{Contracts Q\&A Streamlit application} 
\label{fig:interface}
\end{figure}

The system was evaluated through experiments conducted with two IT contract specialists from BNDES, who validated its performance using a set of 75 contracts. These contracts, including both PDFs and associated metadata, were processed to assess the system’s ability to retrieve relevant information from both unstructured documents and structured data. To evaluate the system's effectiveness in answering various query types, a set of benchmark questions was developed, divided into two categories: direct and indirect questions.

Direct questions refer to those that could be answered using information directly available in the contract PDFs and their metadata. Examples include questions about contract subjects, suppliers, managers, and contract terms. The results demonstrated that for these direct questions, the system consistently provided complete and relevant responses, meeting the users’ expectations for accuracy and comprehensiveness.

Indirect questions, however, required information that would yield better relevance when retrieved from the database. Examples include questions about the number of active contracts, upcoming contract expirations, and specific details regarding exemptions from tender processes. The results for these indirect questions were generally satisfactory, although in certain cases, such as questions about contract inflexibility and exemptions, the answers provided were marked as incomplete. This is likely due to the more complex semantics of the terms involved. For example, the term "Waiver of Bidding" proved challenging for the system, as its meaning was not fully captured in the retrieval process. Adjustments to the prompts or query structure are expected to improve the system’s ability to interpret and respond accurately to these nuanced questions.

User feedback highlighted that one of the system's most valuable features is its ability to seamlessly integrate information from both the structured data store and the unstructured text in contracts. This feature significantly reduces the time users spend locating and accessing relevant contract data, as they would typically need to identify the contracts, open the PDFs, and manually search for information. For instance, the system efficiently retrieves answers regarding contract managers and outlines any penalties related to contractual non-compliance, eliminating the need for users to sift through lengthy documents. By directly addressing questions with specific details, the system enhances the user experience, providing critical information quickly and effectively.

Additionally, users appreciated the system’s capacity to automatically generate visual summaries through its Plotly agent when a table of values was included in the response. This feature was positively received, as it not only provides immediate visual insights but also supports users in preparing professional presentations. By integrating dynamic graph generation directly into the response process, the system offers users a more comprehensive analytical experience, enabling clearer communication and a deeper understanding of contract-related data.

\begin{figure}[ht]
\centering
\fbox{\includegraphics[height=.7\linewidth,width=.7\linewidth]{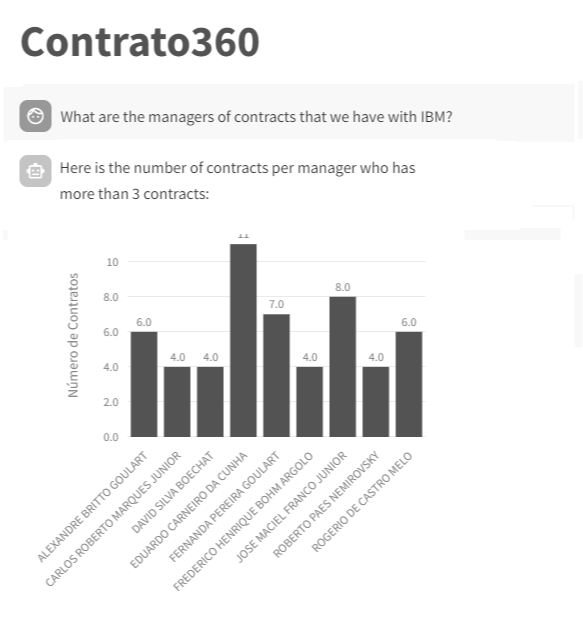}}
\caption{Plotly Agent} 
\label{fig:interface}
\end{figure}

\section{Conclusions and Future Work}
In this work, we presented a comprehensive multi-source question-answer system that integrates unstructured text from contract documents with structured data from relational databases. By employing a combination of Retrieval-Augmented Generation (RAG), Text-to-SQL techniques, and dynamic prompt engineering, we demonstrated how our system efficiently retrieves relevant information from diverse data sources to provide precise and contextually accurate responses. The use of backend agents, particularly the Router Agent, allowed for a flexible and adaptive workflow where queries are dynamically routed to the appropriate processing module—whether that be the RAG agent for text-based retrieval or the SQL agent for direct database queries.

\begin{figure}[ht]
\centering
\fbox{\includegraphics[height=.8\linewidth,width=.8\linewidth]{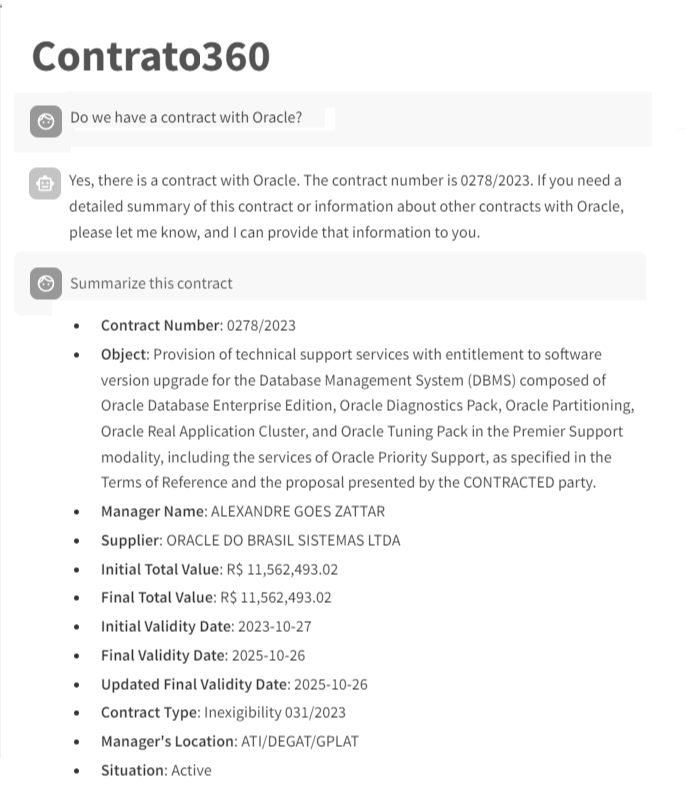}}
\caption{Contract Summarization} 
\label{fig:interface}
\end{figure}

The \ref{fig:interface} demonstrates the ability of Contrato360 in retrieving and summarizing contract information related to Oracle through a question-and-answer interface. Our implementation, which includes the use of ChromaDB as the vectorstore for storing document embeddings and SQLite for managing contract data, ensures that the system can handle complex legal documents while maintaining real-time performance in answering user queries. The combination of these technologies enables the system to provide a seamless experience where both structured and unstructured data are processed cohesively, offering a unified approach to contract management and information retrieval.

Despite the success of our approach, there remain several areas for future development. One significant avenue for improvement is the further refinement of the Router Agent. Currently, it relies on predefined regular expressions to route queries, but integrating machine learning models to dynamically adapt and learn from query patterns could increase the precision and flexibility of the system. Additionally, expanding the system's capability to handle a wider variety of legal documents and domains, beyond contract management, would provide greater scalability and versatility.

Another important direction for future work involves improving the system’s interaction with graph-based data. We have already implemented a Graph Agent to visualize data using bar graphs, but incorporating more advanced data visualizations, such as time-series analysis or multi-dimensional comparisons, would provide users with deeper insights into the retrieved data. Moreover, enhancing the chunking strategy for document segmentation and metadata generation could mitigate the issue of misalignment between query intent and retrieved text, especially for more complex and ambiguous legal queries.

Finally, while our current system integrates effectively with contract documents and databases, there is potential to expand its multi-source retrieval capabilities by incorporating external data sources such as APIs, web services, or even real-time data streams. This would provide users with even more comprehensive and up-to-date information.

In conclusion, while our system already demonstrates significant advancements in combining text-based and structured data retrieval for question-answer tasks, the ongoing development of more sophisticated routing, visualization, and data integration techniques will further enhance its capabilities and application across different domains.

\bibliographystyle{apalike}
{\small
\bibliography{contrato360}}

\vspace{2cm}

\section*{Authors}
\noindent {\bf Antony Seabra} is an IT executive at BNDES, the Development Bank of Brazil, where he leads the Data Engineering team. He received his Master’s Degree in Computer Science (Databases) from PUC-Rio, Brazil, in 2017, and he is currently pursuing his PhD in Computer Science at PUC-Rio under the guidance of Prof. Sérgio Lifschitz. His research interests focus on Databases and their integration with Artificial Intelligence and Natural Language Processing.\\

\noindent {\bf Claudio Cavalcante} is a Data Engineer at BNDES with a solid academic background. He is currently pursuing his Master’s Degree in Computing at PUC-Rio, Brazil, under the guidance of Prof. Sérgio Lifschitz. His research interests lie in Artificial Intelligence and Natural Language Processing.\\

\noindent {\bf João Nepomuceno} received his Bachelor’s Degree in Physics from Universidade Federal Fluminense, Brazil, and he is currently pursuing his Bachelor’s Degree in Computer Science at Universidade Federal Fluminense, Brazil. His research interests include Data Engineering, Artificial intelligence and Natural Language Processing.\\

\noindent {\bf Lucas Lago} is currently pursuing his Bachelor’s Degree in Computer Science at Universidade do Estado do Rio de Janeiro, Brazil. His research interests include Artificial Intelligence and Natural Language Processing.\\

\noindent {\bf Nicolaas Ruberg} is a Data Engineer at BNDES with a solid academic background. He holds a Bachelor’s in Computer Science from the Universidade Federal da Paraíba in Brazil. He further enhanced his expertise by earning a Master’s Degree in Distributed Databases from the Universidade Federal do Rio de Janeiro and later a Master’s in Artificial Intelligence from the University of Bologna in Italy. His research interests include Databases, Artificial Intelligence, and Natural Language Processing.\\

\noindent {\bf Sérgio Lifschitz} is an Associate Professor at PUC-Rio with a research emphasis in Databases. He received his Bachelor’s Degree in Electrical Engineering (1986) and Master’s Degree in the same field (1987) from PUC-Rio and completed his PhD in Computer Science at the École Nationale Supérieure des Télécommunications (ENST Paris) in 1994.

\end{document}